\documentclass[10pt,twocolumn,letterpaper]{article}

\usepackage{xcolor}
\usepackage{iccv}
\usepackage{times}
\usepackage{epsfig}
\usepackage{graphicx}
\usepackage{amsmath}
\usepackage{amssymb}
\usepackage{multirow}
\usepackage{colortbl}
\usepackage{booktabs}
\usepackage{eso-pic}

\usepackage[pagebackref=true,breaklinks=true,letterpaper=true,colorlinks,bookmarks=false]{hyperref}

\iccvfinalcopy 

\def\iccvPaperID{10360} 

\ificcvfinal\fi

\definecolor{mygray}{gray}{.88}

\begin{document}

\title{Masked Spiking Transformer}

\author{
    Ziqing Wang\textsuperscript{\rm 1, \rm 2}\thanks{Equal contribution.} ~~ 
    Yuetong Fang\textsuperscript{\rm 1}\footnotemark[1] ~~
    Jiahang Cao\textsuperscript{\rm 1} ~~
    Qiang Zhang\textsuperscript{\rm 1}~~
    Zhongrui Wang\textsuperscript{\rm 3}\thanks{Corresponding author:  renjingxu@ust.hk\textsuperscript{\rm 1} ~~ zrwang@eee.hku.hk\textsuperscript{\rm 3}.}~~
    Renjing Xu\textsuperscript{\rm 1}\footnotemark[2]~~\\
    \textsuperscript{\rm 1}\small{The Hong Kong University of Science and Technology (Guangzhou)}\\
    \textsuperscript{\rm 2}\small{North Carolina State University} \\
    \textsuperscript{\rm 3}\small{The University of Hong Kong}
}

\maketitle
\ificcvfinal\thispagestyle{empty}\fi

\begin{abstract}
The combination of Spiking Neural Networks (SNNs) and Transformers has attracted significant attention due to their potential for high energy efficiency and high-performance nature. However, existing works on this topic typically rely on direct training, which can lead to suboptimal performance. To address this issue, we propose to leverage the benefits of the ANN-to-SNN conversion method to combine SNNs and Transformers, resulting in significantly improved performance over existing state-of-the-art SNN models. Furthermore, inspired by the quantal synaptic failures observed in the nervous system, which reduces the number of spikes transmitted across synapses, we introduce a novel Masked Spiking Transformer (MST) framework that incorporates a Random Spike Masking (RSM) method to prune redundant spikes and reduce energy consumption without sacrificing performance. Our experimental results demonstrate that the proposed MST model achieves a significant reduction of 26.8\% in power consumption when the masking ratio is 75\% while maintaining the same level of performance as the unmasked model. 

\end{abstract}


\begin{figure}
\begin{center}
\includegraphics[scale=0.49]
{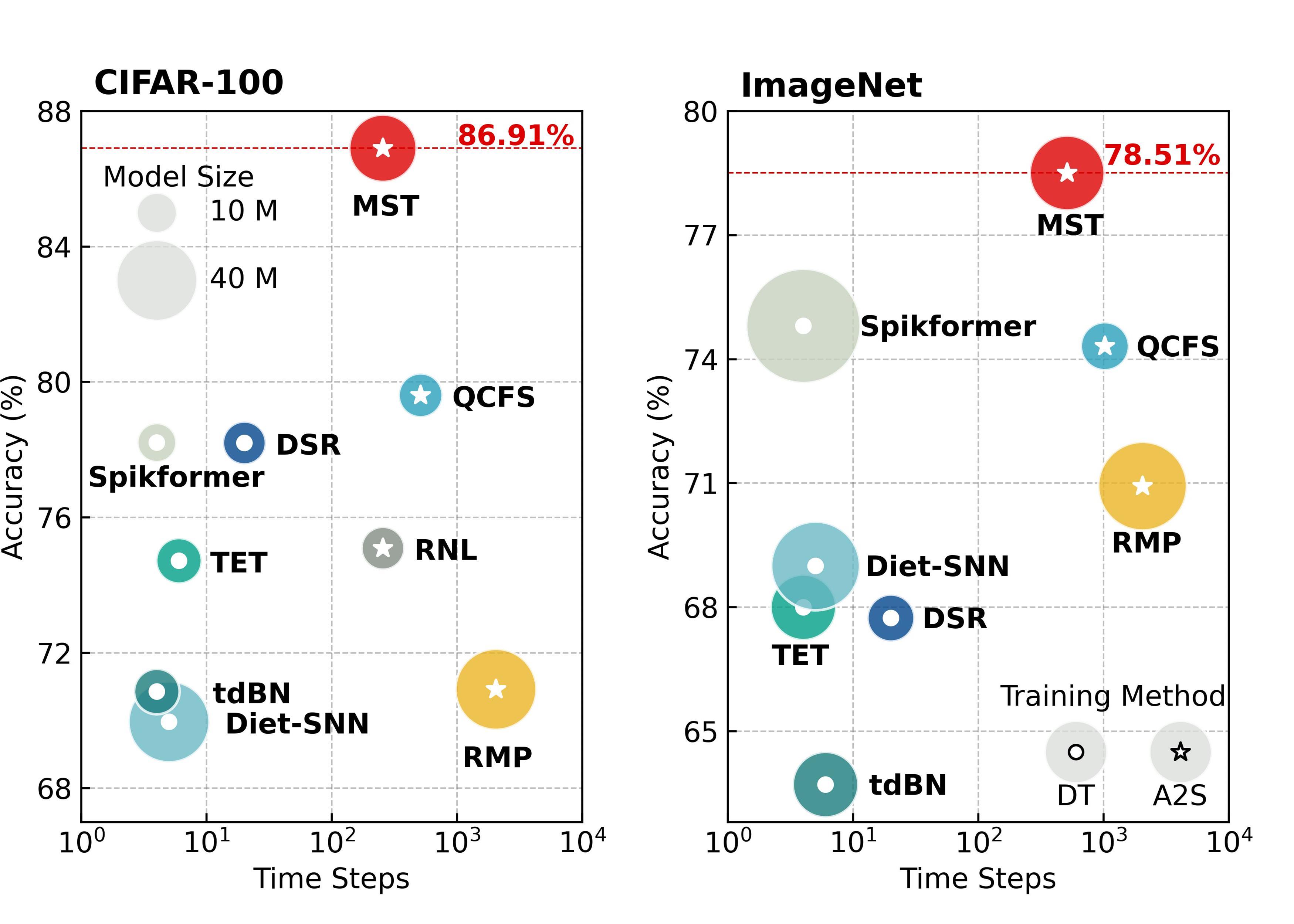}
\end{center}
   \vspace{-0.2cm}
   \caption{Performance of Masked Spiking Transformer (MST) and other state-of-the-art (SOTA) SNN models regarding top-1 accuracy and time steps. The markers, represented by circles and star shapes, denote the direct training (DT) and the ANN-to-SNN conversion method, respectively, where the marker size corresponds to the model size. Results show that the proposed MST model achieves higher accuracy compared to other SNN models.}
   \vspace{-0.5cm}
\label{acc}
\end{figure}
   \vspace{-0.2cm}
\section{Introduction}

Spiking neural networks (SNNs), considered as the next generation neural networks~\cite{maassNetworksSpikingNeurons1997}, are brain-inspired neural networks based on the dynamic characteristics of biological neurons~\cite{mccullochLogicalCalculusIdeas1943, izhikevichSimpleModelSpiking2003}. SNNs have attracted significant attention due to their unique properties in handling sparse data, which can yield great energy efficiency benefits on neuromorphic hardware. Due to their specialties, they have been widely utilized in various fields, such as classification~\cite{mengTrainingHighPerformanceLowLatency2022, kimOptimizingDeeperSpiking2021}, object detection~\cite{caoSpikingDeepConvolutional2015} and tracking~\cite{yangDashNetHybridArtificial2019}, etc. Nevertheless, SNNs currently can hardly realize a comparable performance to that of artificial neural networks (ANNs), especially for complex tasks such as ImageNet~\cite{rathiEnablingDeepSpiking2020a}. 

In order to improve the performance of SNNs, various training methods have been proposed, broadly categorized as the direct training method and the ANN-to-SNN conversion method. Direct training methods leverage a continuous relaxation of the non-smooth spiking mechanism to enable backpropagation with a surrogate gradient function for handling non-differentiability~\cite{neftciSurrogateGradientLearning2019}, but this can lead to unstable gradient propagation and relatively low accuracy compared to leading ANNs
~\cite{ponghiran2022spiking}. Alternatively, ANN-to-SNN conversion methods convert pre-trained ANNs into SNNs for better performance while requiring more time steps, with increased power consumption to reduce conversion errors~\cite{stocklOptimizedSpikingNeurons2021, liFreeLunchANN2021a, buOptimalANNSNNConversion2021a, dingOptimalAnnsnnConversion2021}. Our focus is on implementing the ANN-to-SNN conversion method to narrow the performance gap between leading ANNs and SNNs, but the required long time steps pose challenges in reducing energy consumption. Therefore, identifying strategies to decrease power consumption while maintaining excellent performance is crucial.

The biological nervous system offers valuable insights for addressing the challenges of implementing high-performance Spiking Transformers using the ANN-to-SNN conversion method. The quantal synaptic failure theory suggests that missing information during neuronal signal transmission may not impact the computational information transmitted to a postsynaptic neuron under certain conditions, but can reduce energy consumption and heat production~\cite{levyEnergyEfficientNeuronalComputation2002}. Likewise, in the ANN-to-SNN conversion process, missing spikes can possibly be compensated for by leveraging the correlations between signals in the space and time domains during the information propagation over multiple time steps. Furthermore, neural network models possess lots of redundant connections: prior works reveal that the redundancy in the self-attention module of Transformers can be pruned without significantly impacting performance~\cite{michelAreSixteenHeads2019, voitaAnalyzingMultiheadSelfattention2019}. Therefore, eliminating redundant information during the transmission of neuronal signals can possibly reduce overall energy consumption in the Spiking Transformer model while preserving high performance.

In our work, we propose a Masked Spiking Transformer (MST), which incorporates a Random Spike Masking (RSM) method designed specifically for SNNs. The RSM method randomly selects only a subset of input spikes, significantly reducing the number of spikes involved in the computation process. We evaluate the MST model on both static and neuromorphic datasets, demonstrating its superiority over existing SNN models. Our experiments show that the RSM method can reduce energy consumption on the self-attention module and the MLP module in Transformer, enabling the SNNs to take advantage of energy efficiency and high performance. Furthermore, the proposed RSM method is not limited to Transformer, but can be extended to other backbones such as ResNet and VGG, highlighting its potential as a general technique to improve SNN efficiency. Our results demonstrate the potential of this approach to provide a new direction for developing high-performance and energy-efficient SNN models.

\begin{figure}
\begin{center}
\includegraphics[scale=0.13]
{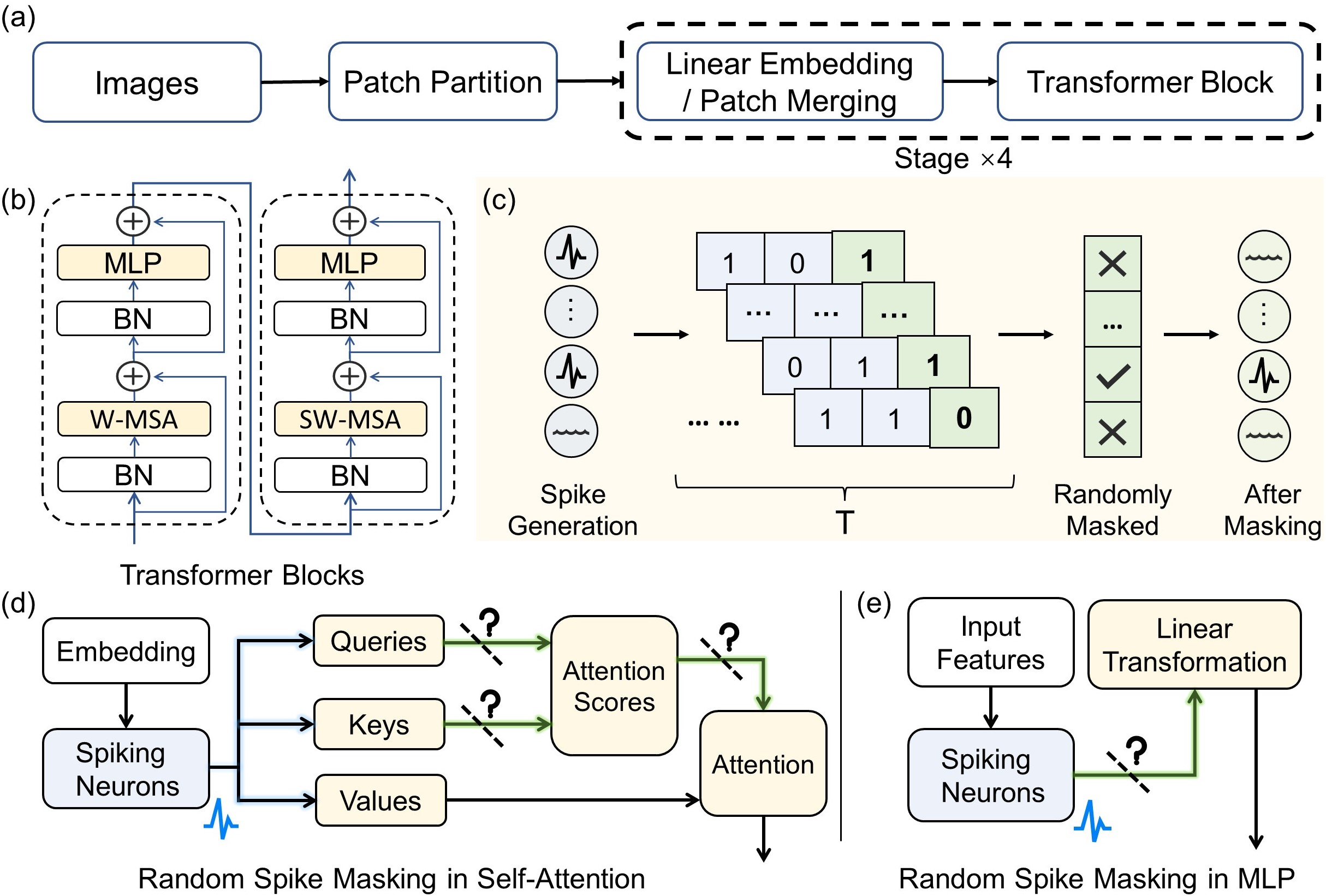}  
\end{center}
\caption{Overview of our Masked Spiking Transformer (MST). (a) Schematic of the model architecture of the Swin Transformer, which is the backbone of our model. (b) Schematic of the proposed Transformer blocks, where BN layers replace the original LN layers. (c) Conceptual illustration of the Random Spike Masking (RSM) method, which involves randomly masking the input spike. (d-e) The RSM method in self-attention and MLP module.}
\label{overview}
\end{figure}

The main contributions of this paper can be summarized as follows:
\begin{itemize}
    \item We propose a Masked Spiking Transformer (MST) using the ANN-to-SNN conversion method. To the best of our knowledge, it is the first exploration of applying the self-attention mechanism fully in SNNs utilizing the ANN-to-SNN conversion method.
    \item The MST model is evaluated on both static and neuromorphic datasets, and the results show that it outperforms SOTA SNNs on all datasets. In specific, the top-1 accuracy of the MST model is 1.21\%, 7.3\%, and 3.7\% higher than the current SOTA SNN model on the CIFAR-10, CIFAR-100, and ImageNet datasets, respectively.
    \item We design a Random Spike Masking (RSM) method for SNNs trained with the ANN-to-SNN conversion method to prune the redundant spikes during inference and save energy consumption.
    \item Extensive experiments show that our proposed RSM is a versatile and general method that can be utilized in other spike-based deep networks, such as ResNet and VGG SNN model variants.
\end{itemize}


\section{Related Work}

\paragraph{Spiking Neural Networks} 
SNNs have gained popularity in the field of brain-inspired intelligence due to their compatibility with neuromorphic hardware and biological properties. With the increasing interest in larger-scale and higher-performance SNNs, recent research has focused on developing novel training algorithms and architectures. Zheng et al. proposed a threshold-dependent batch normalization (tdBN) method based on spatiotemporal backpropagation to train a large-scale SNN model with 50 layers~\cite{zhengGoingDeeperDirectlytrained2021}. Besides, Fang et al. proposed the SEW ResNet architecture for residual learning in deep SNNs to overcome the gradient vanishing problem~\cite{fangDeepResidualLearning2021}. 
Later, they introduced a training algorithm that learns the threshold of each spiking neuron to improve the performance of SNNs~\cite{fangIncorporatingLearnableMembrane2021}. However, these methods mainly discuss the SNN models that are dominated by convolutional layers, such as VGG~\cite{simonyanVeryDeepConvolutional2015a} and ResNet~\cite{heDeepResidualLearning2016} SNN variants. Despite their improvements, the performance of these methods still struggles to match their ANN counterparts, limiting the application of SNNs. In this context, our proposed work focuses on implementing the self-attention mechanism in SNNs to design a Spiking Transformer that improves the performance of SNNs.
\vspace{-0.2cm}
\paragraph{Transformer}
Transformer~\cite{vaswaniAttentionAllYou2017} was first introduced in Natural Language Processing (NLP) and quickly gained popularity for its remarkable capabilities in capturing long-range dependencies. Its success in NLP has inspired researchers to explore its potential in computer vision. Vision Transformer (ViT)~\cite{dosovitskiyImageWorth16x162020} was the first attempt to apply the Transformer to image classification. ViT has achieved impressive results on various computer vision benchmarks, demonstrating the effectiveness of the self-attention mechanism in image understanding. Following the success of ViT, a series of works~\cite{liuSwinTransformerHierarchical2021, hassaniEscapingBigData2021} proposed improvements to the original ViT architecture. Motivated by the success of Transformers and its variations, this paper proposes a new architecture for SNNs that leverages the capacities of the Transformer and the energy efficiency of SNNs. 
\vspace{-0.2cm}
\paragraph{Spiking Transformer}
The combination of the Transformer and SNNs can achieve better performance, which has been discussed in prior studies, including STNet~\cite{zhangSpikingTransformersEventBased2022a} and Spike-T~\cite{zhangSpikeTransformerMonocular2022}. These models utilized separate branches of SNNs and Transformers for feature extraction, leading to the inability to run independently on neuromorphic hardware and failing to exploit the energy efficiency benefits of SNNs fully. In addition, Mueller et al.~\cite{muellerSpikingTransformerNetworks2021} proposed a Spiking Transformer using the ANN-to-SNN conversion method, but they did not implement the self-attention module in SNNs. The recently proposed Spikformer~\cite{zhouSpikformerWhenSpiking2022} directly trained the Transformer in SNNs, but still struggles to achieve comparable performance to leading ANNs. To address these limitations, we apply the self-attention mechanism fully in SNNs by utilizing the ANN-to-SNN conversion method and propose the RSM method to improve both the performance and energy efficiency of the Spiking Transformer. Our model offers a new direction for developing high-performance SNNs using the ANN-to-SNN conversion method.

\section{Methods}


\subsection{Spiking Neuron Model} \label{Spiking Neuron}

For ANNs, the input $\boldsymbol{a}^{l-1}$ to layer $l$ is mapped to the output $\boldsymbol{a}^l$ by a linear transformation matrix $\boldsymbol{W}^l$ and a nonlinear activation function $f(\cdot)$, that is $(l=1,2,3, \cdots, L)$:
\begin{equation}
\label{eq1}
\boldsymbol{a}^l=f\left(\boldsymbol{W}^l \boldsymbol{a}^{l-1}\right)
\end{equation}
\noindent where $f(\cdot)$ is often set as the $ReLU$ activation function.

In SNNs, the Integrate-and-Fire (IF) spiking neuron model is commonly used in ANN-to-SNN conversion~\cite{liFreeLunchANN2021a, buOptimalANNSNNConversion2021a, dingOptimalAnnsnnConversion2021}. The dynamics of the IF model are described by:
\begin{equation}
\label{eq2}
    \boldsymbol{v}^l(t)=\boldsymbol{v}^l(t-1)+\boldsymbol{W}^l \theta^{l-1} \boldsymbol{s}^{l-1}(t)-\theta^l \boldsymbol{s}^l(t)
\end{equation}
\noindent where $\boldsymbol{v}^l(t)$ denotes the membrane potential of neurons in layer $l$ at time-step $t$, which is corresponding to the linear transformation matrix $\boldsymbol{W}^l$, the threshold $\theta^l$, and the binary output spikes of neurons in the previous layer $l-1$, denoted as $\boldsymbol{s}^{l-1}(t)$. The $s^l(t)$  is defined as follows:

\begin{equation}
\label{eq3}
\boldsymbol{s}^l(t)=H\left(\boldsymbol{u}^l(t)-\theta^l\right)
\end{equation}

\noindent where $\boldsymbol{u}^l(t)=\boldsymbol{v}^l(t-1)+\boldsymbol{W}^l \theta^{l-1} \boldsymbol{s}^{l-1}(t)$ denotes the membrane potential of neurons before the trigger of a spike at time-step $t$, $H(\cdot)$ denotes the Heaviside step function. The neurons generate output spikes whenever the membrane potential $\boldsymbol{u}^l(t)$ exceeds the threshold value $\theta^l$, and the membrane potential is reset by subtracting the threshold value to reduce information loss~\cite{rueckauerConversionContinuousvaluedDeep2017}.

\subsection{ANN-to-SNN conversion} \label{ANN2SNN}

To achieve the ANN-SNN conversion, a relationship is established between the rectified linear unit (ReLU) activation of analog neurons in ANNs and the firing rate or postsynaptic potential of spiking neurons in SNNs. This is obtained by summing Eq.~\ref{eq2} from time step 1 to $T$ dividing $T$ on both sides, resulting in the following equation:
\begin{equation}
\label{eq4}
\frac{\boldsymbol{v}^l(T)-\boldsymbol{v}^l(0)}{T} = \frac{\sum_{t=1}^T \boldsymbol{W}^l \theta^{l-1} \boldsymbol{s}^{l-1}(t)}{T} - \frac{\sum_{t=1}^T \theta^l \boldsymbol{s}^l(t)}{T}
\end{equation}
The linear relationship between $\phi^l(T)$ and $\phi^{l-1}(T)$ is established by defining $\phi^l(T)=\frac{\sum_{t=1}^T \theta^l s^l(t)}{T}$ as the average postsynaptic potential:
\begin{equation}
\label{eq5}
\boldsymbol{\phi}^l(T) = \boldsymbol{W}^l \boldsymbol{\phi}^{l-1}(T) - \frac{\boldsymbol{v}^l(T)-\boldsymbol{v}^l(0)}{T}
\end{equation}
The equivalence between Eq.~\ref{eq1} and~\ref{eq5} holds only as T goes to infinity, resulting in a conversion error. To address this issue, we replace the ReLU activation function with the quantization clip-floor-shift (QCFS)~\cite{buOptimalANNSNNConversion2021a} function in the ANNs. 

\subsection{Model Architecture} \label{model}

An overview of the MST is depicted in Fig.~\ref{overview}, where the Swin Transformer~\cite{liuSwinTransformerHierarchical2021} is adopted as the backbone network. To convert the original network into a fully-spiking manner, we incorporate QCFS activation functions after each linear or regularization layer during the training phase, which are replaced with Integrate-and-Fire (IF) neurons in the inference process, resulting in more efficient computation.

The entire computation process in the spiking self-attention module can be formulated as:
\begin{equation}
\begin{split}
Q_{spk}[t] = \operatorname{IF}(X[t]*W_{q}) \\
K_{spk}[t] = \operatorname{IF}(X[t]*W_{k}) \\
\end{split}
\end{equation}
\noindent where $Q_{spk}$, $K_{spk}$ denote the spike matrices of the query and key at $t$ time step, $\operatorname{IF}(\cdot)$ is the IF neuron function, $W_{q}$, $W_{k}$ denote the corresponding weight matrices.
Attention score is defined as:
\begin{equation}
A_{spk}[t] = \operatorname{IF}(\frac{Q_{spk}[t] * K_{spk}^{T}[t]}{\sqrt{d}})
\end{equation}
\noindent where $A_{spk}$ represents the spike matrix of the attention score calculated by the dot product of the query spike matrix and the key spike matrix, and $d$ is a scaling factor equal to the feature dimension of a given attention head. 

The calculation of LN and BN can be expressed by:
\begin{equation}
y=\frac{x-\mathrm{E}[x]}{\sqrt{\operatorname{Var}[x]+\epsilon}} * \gamma+\beta
\end{equation}
where $x$ denotes the input tensor to be normalized, $\mathrm{E}[x]$ and $\operatorname{Var}[x]$ represent the mean and variance of $x$, $\epsilon$ is a small constant added to the variance for numerical stability, $\gamma$ and $\beta$ are trainable scaling and bias parameters respectively, and $y$ is the normalized output of the layer.

In LN, the mean and variance are calculated across the features of each sample in a batch. Therefore, each sample in the batch has its normalization parameters. On the other hand, BN calculates the mean and variance across all samples in a batch for each feature, which means the normalization parameters are shared across all samples in a batch.

\begin{figure}
\begin{center}
\includegraphics[scale=0.2]
{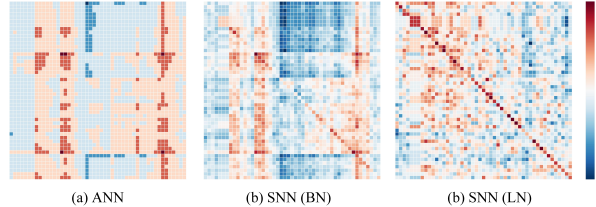}
\end{center}
   \vspace{-0.1cm}
   \caption{Illustration of distributions of (a) post-activation distribution in ANN, and (b-c) cumulative membrane potential distributions of SNN model with BN and LN, respectively. The heatmap shows a similar distribution between ANN and SNN(BN) model, but the distribution between ANN and SNN(LN) is quite different, which leads to performance degradation.}
   \vspace{-0.3cm}
\label{heatmap}
\end{figure}

Normalization is important for ensuring the feasibility of ANN-to-SNN conversion. As illustrated in Fig.~\ref{heatmap}, different normalization approach results in vastly different membrane potential in the inference process. Normalizing along the channel dimension (LN) cause a distribution mismatch between ANN and SNN, which leads to performance degradation while normalizing along the batch dimension (BN) preserves a similar result. Replace all LN layers with BN layers is a straightforward approach to normalize the post-activation distribution of ANNs during the conversion process, but for large datasets like ImageNet, there are convergence issues. We resolve this problem by simply adding a BN layer after each linear layer in the MLP module, inspired by ~\cite{yaoLeveragingBatchNormalization2021}. Mathematically, 
the first layer of the modified MLP module is formulated as follows:
\begin{equation}
\mathrm{MLP}(x) = \operatorname{IF}({\mathrm{Linear}(\mathrm{BN}(x))})
\end{equation}
\noindent where $x$ represents the input tensor, $\mathrm{Linear}$ denotes the linear layer, $\mathrm{BN}$ represents the batch normalization layer. Notably, the MLP module consists of two linear layers, each followed by a BN layer and an IF neuron function.


\subsection{Random Spike Masking Method} \label{SP}

Implementing the Transformer in SNNs using the ANN-to-SNN conversion method presents a challenge due to the high power consumption demand. To address this issue, we propose the Random Spike Masking (RSM) method for reducing redundant spikes during inference.

Traditional ANN pruning methods remove weights with low magnitudes, which are believed to have little impact on the final output of the model~\cite{zhuPruneNotPrune2017a}. In contrast, our RSM method prunes input spikes randomly according to the masking ratio, where each spike ($s = 1$) has the probability of turning into a failure state ($s = 0$) that can be implemented using binary mask matrices.
It is important to note that our proposed RSM method for spike pruning in SNNs differs from Dropout~\cite{srivastavaDropoutSimpleWay2014} used in ANNs. Dropout is a regularization technique that randomly masks some neurons during training, but all neurons are used for inference, and the output needs to be scaled by the retained probability used during training. This means that Dropout cannot directly reduce the number of spikes in the network. Unlike Dropout, the RSM method reduces the computational cost of SNNs by randomly pruning input spikes in both training and inference, explicitly removing redundancy in input spikes while obtaining excellent performance.

The RSM method, illustrated in Fig.~\ref{overview}(c-e), randomly generates a binary mask with the same shape as the input spike matrix based on the masking ratio. This determines which spikes are transmitted for calculation in subsequent neurons, enabling high matrix sparsity. With a masking ratio of 50\%, for example, each spike has a 50\% chance of being masked, reducing the number of spikes and power consumption. Hence, combining both the SNN-based self-attention mechanism with the RSM approach promises a balance between performance and energy efficiency.

\begin{table*}[!t]
\setlength\tabcolsep{9pt} 
\vskip 0.1in
\begin{center}
\begin{small}
\begin{tabular}{ccccccc}
\toprule \toprule
\textbf{Dataset}                  & \textbf{Model}      & \textbf{Method}  & \textbf{Achitecture}     & \textbf{\# Param (M)} & \textbf{Time Steps} & \textbf{Accuracy (\%)} \\
 \cline{1-7}
\multirow{12}{*}{CIFAR10}  & ANN~\cite{liuSwinTransformerHierarchical2021}                       & Direct Training                 & Swin-T   (BN)           & 27.6                  & /                    & 98.14           \\
                            \cline{2-7}
                           & Diet-SNN~\cite{rathiDIETSNNDirectInput2020a}                   & Direct Training                 & VGG-16                  & 39.9                  & 10                   & 93.44           \\
                           & tdBN~\cite{zhengGoingDeeperDirectlytrained2021}                      & Direct Training                 & ResNet-19               & 12.6                  & 4                    & 92.92           \\
                           & TET~\cite{dengTemporalEfficientTraining2022}                       & Direct Training                 & ResNet-19               & 12.6                  & 6                    & 94.50           \\
                           & DSR~\cite{mengTrainingHighPerformanceLowLatency2022}                        & Direct Training                 & PreActResNet-18         & 11.2                  & 20                   & 95.40           \\
                           & Spikformer~\cite{zhouSpikformerWhenSpiking2022}                  & Direct Training                 & Spikformer-4-384         & 9.3                   & 4                    & 95.51           \\
                           \cline{2-7}
                           & RMP~\cite{hanRmpsnnResidualMembrane2020a}                       & ANN-to-SNN                    & VGG-16                  & 39.9                  & 2048                 & 93.63           \\
                           & RNL~\cite{dingOptimalAnnsnnConversion2021}                       & ANN-to-SNN                    & PreActResNet-18         & 11.2                  & 256                  & 93.45           \\
                           & QCFS~\cite{buOptimalANNSNNConversion2021a}                      & ANN-to-SNN                   & RestNet-18              & 11.7                  & 512                  & 96.06           \\
                           \cline{2-7}
                           & \multirow{3}{*}{MST (ours)}       & \multirow{3}{*}{ANN-to-SNN  } & \multirow{3}{*}{Swin-T (BN)}       & \multirow{3}{*}{27.6} &   64                    & \textbf{96.32}       \\
                           &                                     &                          &                                    &                       &    128                  & \textbf{97.06}           \\
                           &                                      &                          &                                      &                       &      256 & \textbf{97.27}         \\
 \cline{1-7}
\multirow{12}{*}{CIFAR100} & ANN~\cite{liuSwinTransformerHierarchical2021}                       & Direct Training                 & Swin-T (BN)             & 27.6                  & /                    & 88.72           \\ \cline{2-7}
                           & Diet-SNN~\cite{rathiDIETSNNDirectInput2020a}                  & Direct Training                & VGG-16                  & 39.9                  & 5                    & 69.67           \\
                           & tdBN~\cite{zhengGoingDeeperDirectlytrained2021}                       & Direct Training                & ResNet-19               & 12.6                  & 4                    & 70.86           \\
                           & TET~\cite{dengTemporalEfficientTraining2022}                       & Direct Training                & ResNet-19               & 12.6                  & 6                    & 74.72           \\
                           & DSR~\cite{mengTrainingHighPerformanceLowLatency2022}                       & Direct Training                 & PreActResNet-18         & 11.2                  & 20                   & 78.50           \\
                           & Spikformer                 & Direct Training                 & Spikformer-4-384         & 9.3                   & 4                    & 78.21           \\
                           \cline{2-7}
                           & RMP~\cite{hanRmpsnnResidualMembrane2020a}                         & ANN-to-SNN                    & VGG-16                  & 39.9                  & 2048                 & 70.93           \\
                           & RNL~\cite{dingOptimalAnnsnnConversion2021}                       & ANN-to-SNN                   & PreActResNet-18         & 11.2                  & 256                  & 75.10           \\
                           & QCFS~\cite{buOptimalANNSNNConversion2021a}                       & ANN-to-SNN                    & RestNet-18              & 11.7                  & 512                  & 79.61           \\
                           \cline{2-7}
                           & \multirow{3}{*}{MST (ours)}       & \multirow{3}{*}{ANN-to-SNN  } & \multirow{3}{*}{ Swin-T   (BN)}      & \multirow{3}{*}{27.6} &    64                   & \textbf{85.40}       \\
                           &                            &                          &                                              &                       &  128                    & \textbf{86.73}           \\
                           &                            &                          &                                                &                       &        256                  & \textbf{86.91}       \\
 \cline{1-7}
\multirow{10}{*}{ImageNet} & ANN~\cite{liuSwinTransformerHierarchical2021}                       & Direct Training                 & Swin-T   (BN)           & 28.5                  & /                    & 80.51           \\
\cline{2-7}
                           & Diet-SNN~\cite{rathiDIETSNNDirectInput2020a}                  & Direct Training                 & VGG-16                  & 39.9                  & 5                    & 69.00           \\
                         & tdBN~\cite{zhengGoingDeeperDirectlytrained2021}                       & Direct Training                & SEW-ResNet-34               & 21.8                  & 4                    & 67.04           \\
                           & TET~\cite{dengTemporalEfficientTraining2022}                        & Direct Training                 & SEW-ResNet-34           & 21.8                  & 4                    & 68.00           \\
                           & DSR~\cite{mengTrainingHighPerformanceLowLatency2022}                       & Direct Training               & PreActResNet-18         & 11.2                  & 50                   & 67.74           \\
                           & Spikformer ~\cite{zhouSpikformerWhenSpiking2022}               & Direct Training                 & Spikformer-8-768         & 66.3                  & 4                    & 74.81           \\
                           \cline{2-7}
                           & RMP~\cite{hanRmpsnnResidualMembrane2020a}                       & ANN-to-SNN                  & VGG-16                  & 39.9                  & 2048                 & 73.09           \\
                           & QCFS~\cite{buOptimalANNSNNConversion2021a}                      & ANN-to-SNN                    & RestNet-18              & 11.7                  & 1024                 & 74.32           \\
                           \cline{2-7}
                           & \multirow{3}{*}{MST (ours)} & \multirow{3}{*}{ANN-to-SNN} & \multirow{3}{*}{ Swin-T   (BN)}      & \multirow{3}{*}{28.5} & 128                    & \textbf{77.88}         \\
                           &                            &                          &                                   &                       &  256                   & \textbf{78.37}           \\
                           &                            &                          &                                  &                       &   512 & \textbf{78.51}        \\ \hline     
\bottomrule
\end{tabular}
\end{small}
\end{center}
\vskip -0.1in
\caption{Performance comparison between the proposed model and the SOTA models on different static datasets.}
\label{table1}
\end{table*}

\section{Experiments}

We conduct extensive experiments on both static datasets including CIFAR-10~\cite{lecunGradientbasedLearningApplied1998a}, CIFAR-100~\cite{krizhevskyLearningMultipleLayers2009a}, and ImageNet datasets~\cite{dengImagenetLargescaleHierarchical2009}, and neuromorphic datasets including CIFAR10-DVS~\cite{liCIFAR10DVSEventStreamDataset2017}, N-Caltech101~\cite{orchardConvertingStaticImage2015b}, and N-Cars~\cite{sironiHATSHistogramsAveraged2018a}, ActionRecognition~\cite{miaoNeuromorphicVisionDatasets2019}, and ASL-DVS~\cite{biGraphbasedObjectClassification2019} datasets, to validate the effectiveness of the MST model. In addition, we evaluate the effect of the RSM approach on accuracy and energy efficiency. More details of the training can be found in the supplementary.

\subsection{Performance on Static Datasets}

Tab.~\ref{table1} presents a comprehensive comparison of the MST model with the current SOTA SNN models on the CIFAR-10/100 and ImageNet datasets. The results show that the proposed MST model outperforms all other models in terms of top-1 accuracy on all three datasets.

For low latency inference with time steps less than 20, several models, such as Diet-SNN~\cite{rathiDIETSNNDirectInput2020a}, tdBN~\cite{zhengGoingDeeperDirectlytrained2021}, TET~\cite{dengTemporalEfficientTraining2022}, and DSR~\cite{mengTrainingHighPerformanceLowLatency2022}, adopt the direct training approach, but their top-1 accuracy is relatively low, compared to that of the MST model, in which tdBN~\cite{zhengGoingDeeperDirectlytrained2021}  is  10\% less accurate on the CIFAR-100 and ImageNet datasets. 

For high latency inference with time steps greater than 100, RMP~\cite{hanRmpsnnResidualMembrane2020a}, RNL~\cite{dingOptimalAnnsnnConversion2021}, and QCFS~\cite{buOptimalANNSNNConversion2021a} leverage ANN-to-SNN conversion method. The MST model also employs this method but requires only 64 time steps to achieve superior performance, which is far fewer than RMP ~\cite{hanRmpsnnResidualMembrane2020a}. This suggests that the MST model can achieve the highest accuracy in a reasonable number of time steps.

Compared to transformer-based SNN models such as Spikformer~\cite{zhouSpikformerWhenSpiking2022}, the MST model achieves higher top-1 accuracy on all three datasets. Specifically, the top-1 accuracy of the MST model is 1.8\%, 8.7\%, and 3.7\% higher than Spikformer on the CIFAR-10, CIFAR-100, and ImageNet datasets, respectively. Moreover, the MST model has significantly fewer parameters compared to the 8-layer Spikformer~\cite{zhouSpikformerWhenSpiking2022} model.

\subsection{Performance on Neuromorphic Datasets} \label{performanceonevnet}

\begin{table}[!t]
\setlength\tabcolsep{2pt} 

\begin{center}
\begin{small}
\begin{tabular}{cccc}
\toprule \toprule
\textbf{Dataset}              & \textbf{Model}           & \textbf{Time   Steps}          &                                  \textbf{Accuracy (\%)} \\ \cline{1-4}
\multirow{11}{*}{CIFAR10-DVS}  & Swin-T (BN)           & /                                                                  & 88.98             \\\cline{2-4}
                              & TA-SNN~\cite{yaoTemporalwiseAttentionSpiking2021}        & 10                             & 72.00                \\
                              & PLIF~\cite{fangIncorporatingLearnableMembrane2021}          & 20                               & 74.80             \\
                              & Dspkie~\cite{liDifferentiableSpikeRethinking2021}        & 10                              & 75.40              \\
                              & DSR ~\cite{mengTrainingHighPerformanceLowLatency2022}           & 10                        & 77.30              \\
                              & TET~\cite{dengTemporalEfficientTraining2022}           & 10                             & 83.17             \\
                              & NDA ~\cite{liNeuromorphicDataAugmentation2022}          & 10                               & 81.70              \\
                              & Spikformer ~\cite{zhouSpikformerWhenSpiking2022}   & 10                                & 80.90              \\
                              \cline{2-4}
                              & \multirow{3}{*}{MST (ours)}       & 128                                                & \textbf{86.60}    \\ 
                              &       & 256                                                & \textbf{87.20}    \\ 
                              &       & 512                                                & \textbf{88.12}    \\ 
                              \cline{1-4}
\multirow{6}{*}{N-CALTECH101} & Swin-T (BN)            & /                            & 92.00             \\\cline{2-4}
                              & SALT~\cite{kimOptimizingDeeperSpiking2021a}          & 20                                & 55.00                \\
                              & NDA~\cite{liNeuromorphicDataAugmentation2022}           & 10                                 & 83.70              \\
                                \cline{2-4}
                              & \multirow{3}{*}{MST (ours)}      & 64                          & \textbf{84.71}    \\ 
                              &        & 128                          & \textbf{89.42}    \\ 
                              &        & 256                          & \textbf{91.38}    \\ 
                              \cline{1-4}
\multirow{6}{*}{N-CARS}       & Swin-T (BN)           & /                             & 97.14             \\\cline{2-4}
                              & CarSNN~\cite{vialeCarsnnEfficientSpiking2021}        & 10                           & 86.00                \\
                              & NDA ~\cite{liNeuromorphicDataAugmentation2022}          & 10                               & 91.90              \\
                            \cline{2-4}
                              & \multirow{3}{*}{MST (ours)}      & 32                            & \textbf{94.67}   \\ 
                              &       & 64                            & \textbf{96.58}   \\ 
                              &       & 128                            & \textbf{97.28}   \\ 
                              \cline{1-4}
\multirow{6}{*}{Action Recognition}       & Swin-T (BN)           & /                             & 90.14             \\\cline{2-4}
                              & STCA~\cite{guSTCASpatiotemporalCredit2019}        & 10                           & 71.20                \\
                              & Mb-SNN ~\cite{liuEventbasedActionRecognition2021a}          & 10                               & 78.10             \\
                              \cline{2-4}
                              & \multirow{3}{*}{MST (ours)}       & 64                          & \textbf{84.76}    \\ 
                              &        & 128                          & \textbf{86.92}    \\ 
                              &        & 256                            & \textbf{88.21}   \\ 
                              \cline{1-4}

\multirow{6}{*}{ASL-DVS}       & Swin-T (BN)           & /                             & 99.90            \\\cline{2-4}
                              & Meta-SNN~\cite{stewartMetalearningSpikingNeural2022}        & 100                           &   96.04             \\
                              \cline{2-4}
                              & \multirow{3}{*}{MST (ours)}       & 64                          & \textbf{98.04}    \\ 
                              &     & 128                          & \textbf{98.51}   \\ 
                              &        & 256                            & \textbf{99.10}   \\ 
                              \hline                    \bottomrule  
\end{tabular}
\end{small}
\end{center}
\vskip -0.1in
\caption{Performance comparison between the proposed model and the SOTA models on different neuromorphic datasets.}
\label{tb2}
\end{table}

\begin{figure}[!t]
\begin{center}
\includegraphics[scale=0.17]
{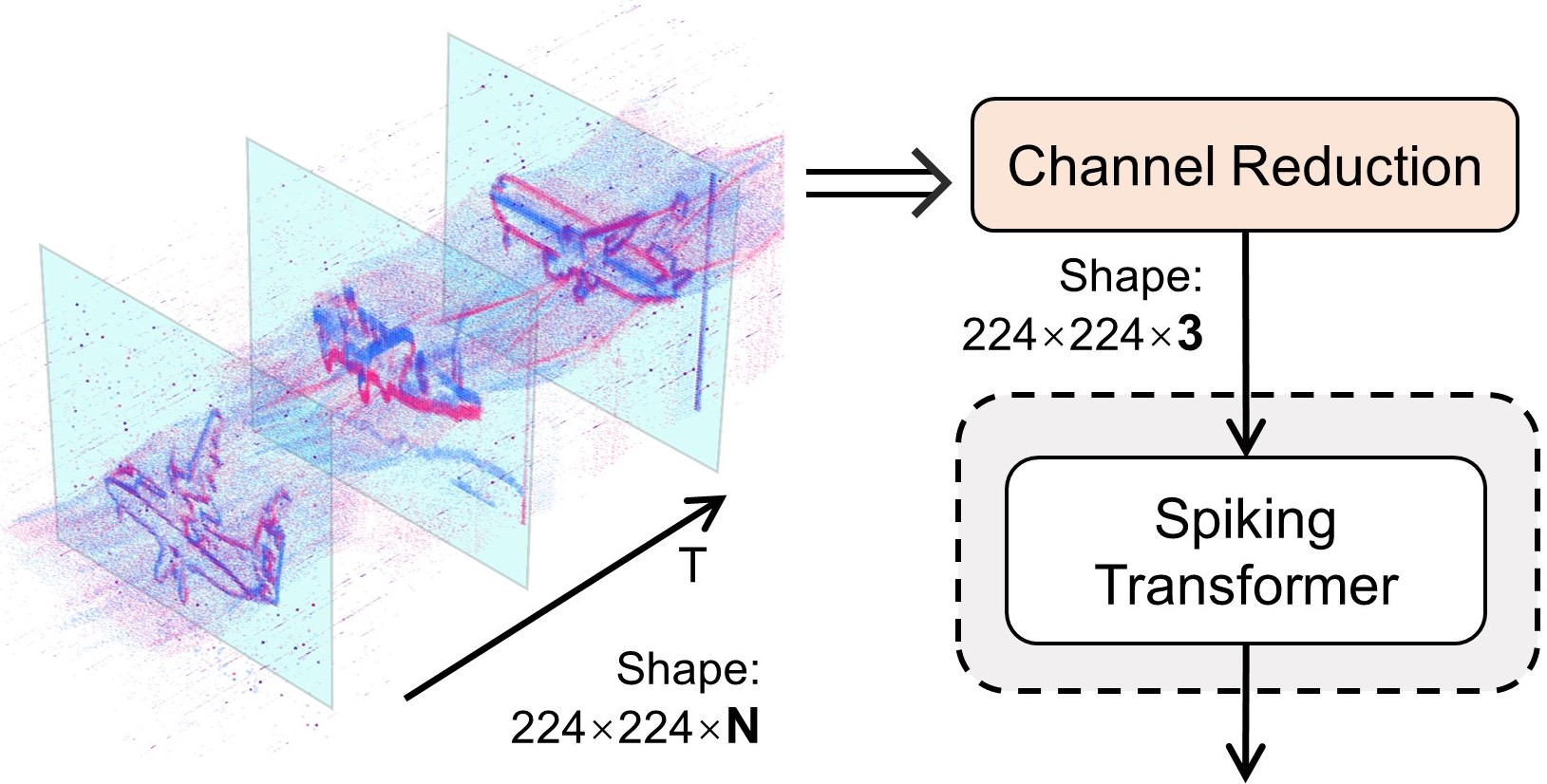}
\end{center}
   \caption{Model Architecture for Neuromorphic Datasets.}
\label{eventmodel}
\end{figure}

We showcase the suitability of the MST model for processing event-based data by evaluating its performance on neuromorphic datasets. Fig.~\ref{eventmodel} shows the framework for the MST model to process the neuromorphic datasets. We utilize a frame-based representation for preprocessing, where the event streams are transformed into a sequence of high-rate frames. Each event in the stream consists of four dimensions, including two spatial coordinates (x, y), timestamp, and polarity. The frames are integrated into an input tensor of size (n*C, H, W), where n is the number of frames, C is the number of original channels (which equals two, representing polarity), and H and W represent the height and width of the input, respectively. To align the input dimension with the model, an additional reduction layer is added to the first layer of the entire model, which reduces the channel dimension to 3. This enables us to leverage pre-training weights from the ImageNet dataset, accelerating the training convergence. Data augmentation for SNNs~\cite{liNeuromorphicDataAugmentation2022} is also applied to improve the accuracy. The ensuing experimental results demonstrate that the proposed fine-tuning approach for ANN-to-SNN conversion achieves remarkably high accuracy.

The experimental results presented in Tab.~\ref{tb2} demonstrate the effectiveness of the proposed MST model in processing neuromorphic datasets. We compare the MST model with several SOTA SNN models on five popular neuromorphic datasets, including CIFAR10-DVS, N-Caltech101, N-Cars, Action Recognition, and ASL-DVS dataset. The experimental results show that the MST model achieves the highest top-1 accuracy on all datasets.

CIFAR10-DVS, N-Caltech101, and N-Cars datasets are constructed by converting the static datasets into event data by using event-based cameras. Tab.~\ref{tb2} shows that the MST model outperforms other SOTA SNN models significantly, achieving improvements of 4.95\%, 7.68\%, and 5.38\% on CIFAR10-DVS, N-Caltech101, and N-Cars, respectively. The Action Recognition dataset consists of a series of human actions captured by event-based cameras. By adopting the data preprocessing method of~\cite{krishnanBenchmarkingConventionalVision2022}, our MST model achieves a top-1 accuracy of 88.21\%, which is far higher than other models. Additionally, the ASL-DVS dataset is a large 24-class dataset of gestures and the experimental results demonstrate that our MST model outperforms other SOTA models by 3.06\%. 

These results highlight the effectiveness of our proposed MST model in processing various types of neuromorphic datasets.

\subsection{Effectiveness of the Random Spike Masking Method}

The ANN-to-SNN conversion method typically involves a large number of time steps, leading to high power consumption. To address this issue, we propose the RSM method for spike pruning, and we demonstrate its effectiveness in the following section.

Drawing inspiration from knowledge distillation~\cite{hintonDistillingKnowledgeNeural2015}, we use the model without masked as a teacher model, and the model after masked as a student model. By fine-tuning the student model, the masked model achieves high performance comparable to the original model without masking  in terms of accuracy. This finding highlights the efficacy of the knowledge distillation technique in training masked models without compromising their performance. Consequently, we employ the fine-tuning approach as the training method for our subsequent comparative analysis.


\begin{figure}
\begin{center}
\includegraphics[scale=0.28]
{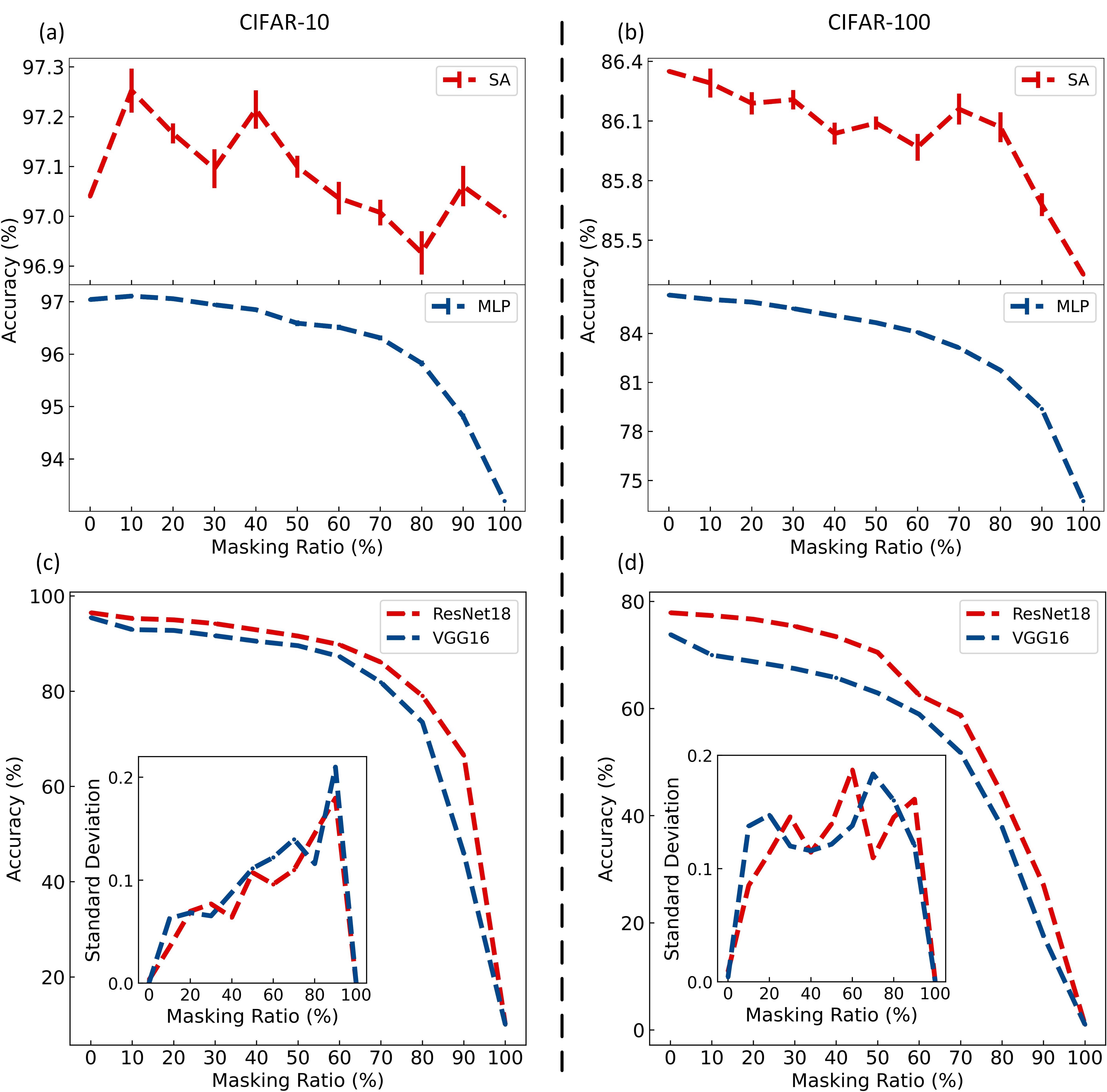}
\end{center}
   \caption{ The effectiveness of the RSM method in (a) SA and (b) MLP modules of the MST model, as well as in Spiking (c) ResNet18 and (d) VGG16 models, with varying masking ratios. The inset photographs show the standard deviation of accuracy in 10 runs.}
   \vspace{-0.3cm}
\label{resneterror}
\end{figure}

We evaluate the RSM method in two critical modules in the Transformer: the self-attention (SA) module and the Multi-layer Perceptron (MLP) module. As shown in Fig.~\ref{overview}(d-e), we apply the RSM method to the Query, Key, and Attention matrices in the SA module, as well as the output spike matrix of the first fully connected layer within each block of the MLP module.

Fig.~\ref{resneterror}(a) and (b) depict the variation in accuracy for different masking ratios on the CIFAR-10 and CIFAR-100 datasets, respectively. The experimental results demonstrate that the accuracy decreases with the increasing masking ratio for both the SA and MLP modules, but their sensitivity to masking ratio changes differs. Specifically, the accuracy of the SA module remains stable over a certain range of masking ratios. In contrast, the accuracy of the MLP module declines more sharply and is more sensitive to masking ratio changes. Consistent with prior research ~\cite{michelAreSixteenHeads2019, voitaAnalyzingMultiheadSelfattention2019}, the redundancy in the SA module enables the MST to withstand the losses caused by missing input spikes without significantly affecting the overall performance of the SNN.

\begin{figure}[!t]
\begin{center}
\includegraphics[scale=0.35]
{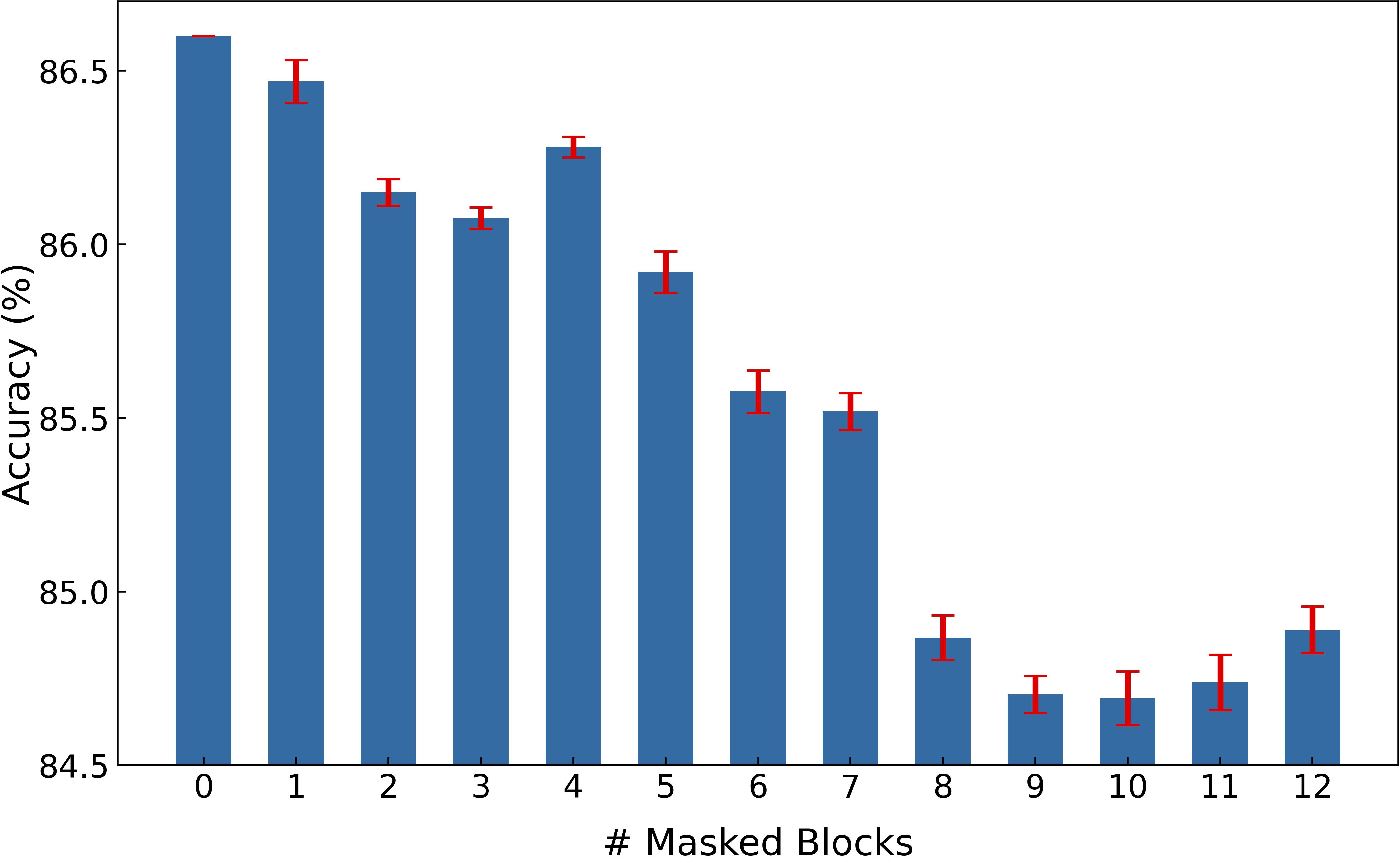}
\end{center}
   \caption{Comparison of the accuracy of masking different numbers of blocks on the CIFAR-100 dataset.}
      \vspace{-0.4cm}
\label{blockcompare}
\end{figure}

In addition to the variation in sensitivity to changes in the masking ratio between different modules, our experiments also demonstrate that the sensitivity can vary depending on the dataset used. The CIFAR-10 dataset is relatively less sensitive to changes in the masking ratio, with only a 0.1\% decrease in accuracy when the masking ratio reaches 80\%. In contrast, the CIFAR-100 dataset is more sensitive to changes in the masking ratio, experiencing a more substantial accuracy decline as the masking ratio increases.

We also investigate the potential of the RSM method as a general approach for reducing power consumption in the context of ANN-to-SNN conversion. By applying the RSM method to other SNN models like Spiking ResNet18 and VGG16 and introducing a masking ratio to each convolution layer, we aim to decrease the transmitted spikes. Our results, presented in Fig.~\ref{resneterror}, demonstrate that the redundancy of these models can also be leveraged to maintain performance while reducing power consumption within a specific range. This indicates the potential of the RSM method for widespread applications in various SNN models for improving the energy efficiency.

Fig.~\ref{blockcompare} illustrates the impact of masking different numbers of blocks on the accuracy of the MST model. y. Our results indicate a non-linear, positive relationship between the number of masked blocks and performance loss. Notably, masking the first 2-5 blocks causes a slight performance loss, and the loss increases with more masked blocks until saturation at around 9-10 blocks. These observations provide valuable insights for designing partially masked blocks.

To better evaluate the effects of the RSM method on energy consumption reduction, we utilize theoretical estimates of energy consumption on neuromorphic chips based on previous studies~\cite{caoSpikingDeepConvolutional2015a, dingOptimalAnnsnnConversion2021}. Assuming that a spike activity consumes $\alpha$ Joules and 1 time-step takes $1 \mathrm{~ms}$. Then the power model is defined as:
\begin{equation}
   P=\frac{\text { total spikes }}{1 \times 10^{-3}} \times \alpha (\text {Watts}) 
\end{equation}
\begin{table}[!t]
\setlength\tabcolsep{3pt} 
\begin{center}
\begin{small}
\begin{tabular}{cccc}
\toprule \toprule
\textbf{Model}            & \textbf{Random   Ratio} & \textbf{P ($\alpha$  Watts)} & \textbf{Accuracy   (\%)} \\
\cline{1-4}
\multirow{3}{*}{MST}      & 0\%                     & 3.9G ($\times$ 1)              & 97.27 (+0)                \\
                          & 50\%                    & 3.2G ($\times$ 0.82)            & 97.25 (-0.02)            \\
                          & 75\%                    & 2.9G ($\times$ 0.74)            & 97.29  (+0.02)            \\\cline{1-4}
\multirow{3}{*}{ResNet18} & 0\%                     & 58.2M ($\times$ 1)              & 96.48 (+0)                \\ 
                          & 50\%                    & 40.7M ($\times$ 0.70)           & 92.88 (-3.6)             \\
                          & 75\%                    & 34.1M  ($\times$ 0.58)          & 82.68 (-13.8)           \\ \cline{1-4}
\multirow{3}{*}{VGG16}    & 0\%                     & 24.4M ($\times$ 1)              & 95.46 (+0)                \\ 
                          & 50\%                    & 18.9M ($\times$ 0.77)           & 89.56 (-5.9)             \\
                          & 75\%                    & 16.7M ($\times$ 0.68)           & 79.09 (-16.37)    \\ \hline                    \bottomrule      
\end{tabular}
\end{small}
\end{center}
\vskip -0.1in
\caption{Comparison of power consumption and accuracy between models with a Masking ratio of 0\%, 50\%, and 75\% on the CIFAR-10 dataset. $(\cdot\cdot\cdot)$ in the table denotes the power consumption/accuracy compared to the unmasked model (with 0\% random ratio).}
\vspace{-0.4cm}
\label{tb_sops}
\end{table}

Tab.~\ref{tb_sops} presents the power consumption and accuracy of the MST, ResNet18, and VGG16 at 0\%, 50\%, and 75\% masking ratios on the CIFAR-10 dataset. The RSM method is applied to the SA module in the MST and each block in ResNet18 and VGG16. The results demonstrate that the RSM method has a direct effect on the number of spikes transmitted, which in turn reduces power consumption. In specific, applying the RSM method to the MST model can reduce power consumption by 26.8\% without any loss in performance when the masking ratio reaches 75\%. This finding suggests that there exist substantial redundant spikes in the SA module that can be pruned for better energy efficiency. As for ResNet18 and VGG16, the RSM method also yields significant power reduction. For instance, ResNet18 can reduce power consumption by 30.1\% with a moderate accuracy loss of 3.6\%. However, excessive masking leads to a significant drop in accuracy.

In summary, our results demonstrate that the proposed RSM method is effective when applied to the SA module in the Transformer, as well as other SNN models.  By eliminating redundant spikes, the RSM method reduces power consumption while preserving performance, making it a promising approach for energy-efficient ANN-to-SNN conversion.

\begin{figure}
\begin{center}
\includegraphics[scale=0.78]
{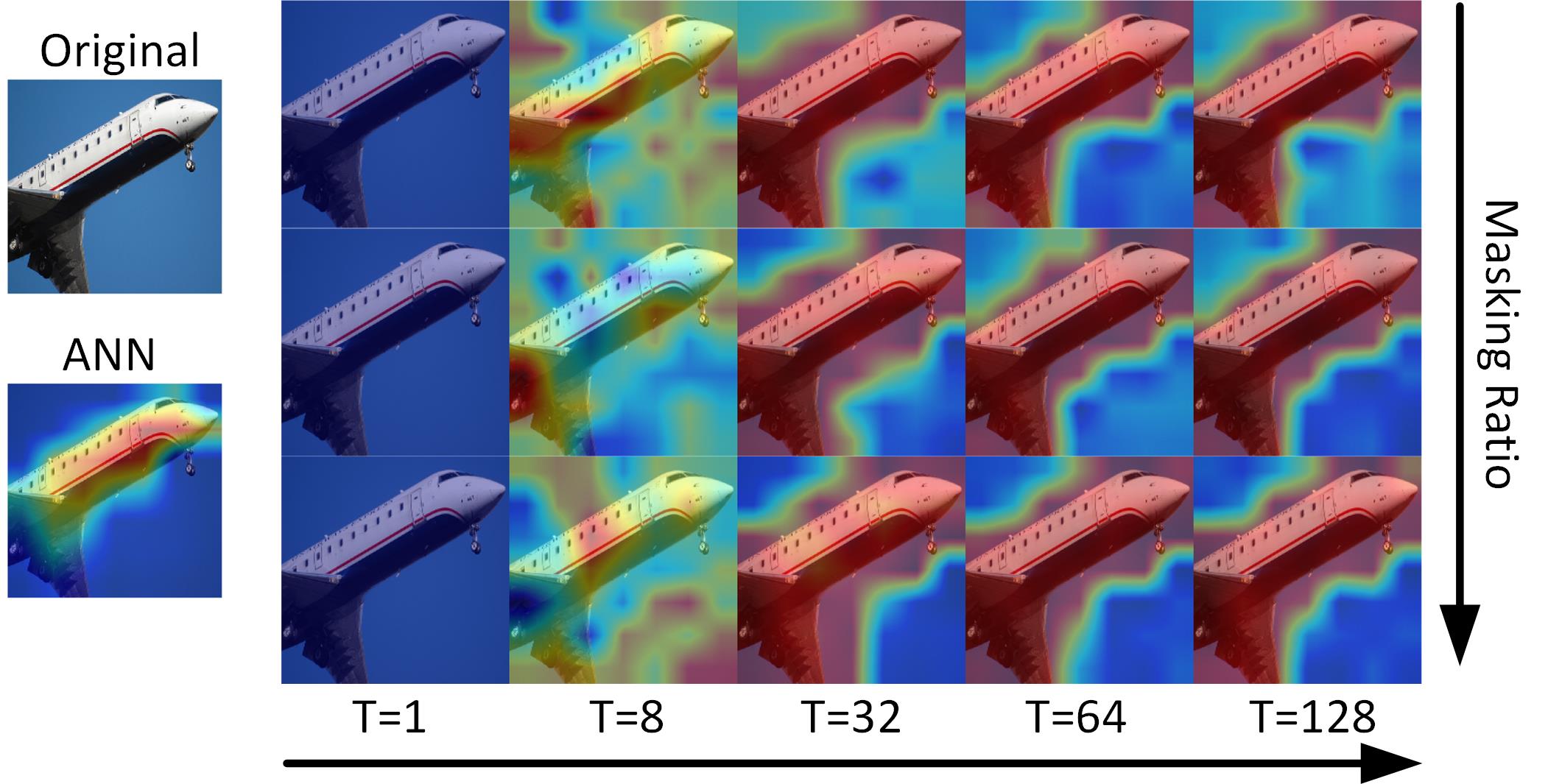}
\end{center}
   \caption{Comparison of attention maps between the MST model with 0\%, 50\%, and 75\% masking ratio (from top to bottom) with 5 different time steps.}
   \vspace{-0.3cm}
\label{attentionmap}
\end{figure}

To visualize the impact of different masking ratios on attention maps, we compared the models with 0\%, 50\%, and 75\% masking ratios using the Spike Activation Maps (SAM) method. The results, shown in Fig.~\ref{attentionmap}, demonstrate that the models with different masking ratios concentrate on similar areas of the object at the same time step, with the red parts outlining the object. Comparison with the ANN model using the ScoreCAM method~\cite{wangScoreCAMScoreWeightedVisual2020} reveals that both models focus on similar key information. These results suggest that the proposed RSM method preserves the regions of interest within the model, contributing to accuracy preservation.

\section{Discussion}


In this paper, we proposed the Masked Spiking Transformer (MST) with the Random Spike Masking (RSM) method. By pruning input spikes, the proposed RSM method effectively reduces power consumption while maintaining the performance of the model within a certain range.

Though our experiments highlight the superiority of the MST model over state-of-the-art SNN models, our model still has limitations that need addressing. A key constraint is the relatively long time steps needed by the ANN-to-SNN conversion method, limiting the suitability of the proposed model for real-time applications with strict timing demands.

Furthermore, even though the RSM method reduces the number of spikes and energy consumption, the experiment in this article merely investigates its applicability of ANN-to-SNN conversion, thus exhibiting relatively high energy consumption compared to direct-trained SNN models, which take fewer timesteps. Consequently, future research could apply the RSM method to direct training methods to optimize energy consumption further and make SNNs more practical for real-world applications. Additionally, adopting different masking ratios across layers may achieve a better balance between performance and energy efficiency, as experiments suggest each layer has a varying effect on overall performance.

\section{Conclusion}

In this work, we propose a Masked Spiking Transformer (MST) framework that combines the energy efficiency of SNNs with the high-performance self-attention mechanism of Transformers using the ANN-to-SNN method. Additionally, we introduce a Random Spike Masking (RSM) method to prune input spikes, thus reducing power consumption. The experimental results demonstrate that the MST model outperforms current SOTA SNN models on both static and neuromorphic datasets. Furthermore, the proposed RSM method shows significant power reduction while maintaining performance in different modules of the Transformer and other SNN models. Our work opens up new possibilities for developing high-performance SNN models, paving the way for future research in this area.



\clearpage

{\small
\bibliographystyle{ieee_fullname}
\bibliography{FST}
}

\end{document}